\documentclass{article}

\usepackage{natbib}
\usepackage[colorlinks=true, citecolor=blue]{hyperref}

\usepackage{svg}
\usepackage{amsmath}
\usepackage{subcaption}
\usepackage{multirow}
\newcommand{\ms}{\!\:}
\usepackage{enumitem}
\usepackage{caption}

\usepackage{authblk}   
\usepackage{setspace}  

\usepackage[margin=3cm, top=2cm, bottom=2cm]{geometry}

\title{Towards Explainable Deep Learning for Ship Trajectory Prediction in Inland Waterways\footnote{
		This is a preprint of a paper published in the Proceedings of the 35th European Safety and Reliability \& the 33rd Society for Risk Analysis Europe Conference. DOI of the published version: 
		\href{https://doi.org/10.3850/978-981-94-3281-3_ESREL-SRA-E2025-P1370-cd}{10.3850/978-981-94-3281-3\_ESREL-SRA-E2025-P1370-cd}. Reproduced here with permission of the publisher. For citation purposes, please refer exclusively to the published version.
	}}

\author[1]{Tom Legel}
\author[2]{Dirk Söffker}
\author[3]{Roland Schätzle}
\author[4]{Kathrin Donandt}

\affil[1]{Department Hydraulic Engineering in Inland Areas\\
	Federal Waterways Engineering and Research Institute, Germany\\
	\texttt{tom.legel@baw.de}}

\affil[2]{Chair of Dynamics and Control\\
	University of Duisburg-Essen, Germany\\
	\texttt{soeffker@uni-due.de}}

\affil[3]{Department of Technology (Computer Science)\\
	Baden-Wuerttemberg Cooperative State University, Germany\\
	\texttt{roland.schaetzle@dhbw-karlsruhe.de}}

\affil[4]{Institute for Sustainable and Autonomous Maritime Systems\\
	University of Duisburg-Essen, Germany\\
	\texttt{kathrin.donandt@uni-due.de}}
\date{}  

\begin{document}

	\maketitle
	\begin{abstract} 
		Accurate predictions of ship trajectories in crowded environments are essential to ensure safety in inland waterways traffic. 
		Recent advances in deep learning promise increased accuracy even for complex scenarios. While the challenge of ship-to-ship awareness is being addressed with growing success, the explainability of these models is often overlooked, potentially obscuring an inaccurate logic and undermining the confidence in their reliability. 
		This study examines an LSTM-based vessel trajectory prediction model by incorporating trained ship domain parameters that provide insight into the attention-based fusion of the interacting vessels’ hidden states. This approach has previously been explored in the field of maritime shipping, yet the variety and complexity of encounters in inland waterways allow for a more profound analysis of the model’s interpretability. 
		The prediction performance of the proposed model variants are evaluated using standard displacement error statistics. Additionally, the  plausibility of the generated ship domain values is analyzed. With an final displacement error of around 40 meters in a 5-minute prediction horizon, the model performs comparably to similar studies. Though the ship-to-ship attention architecture enhances prediction accuracy, the weights assigned to vessels in encounters using the learnt ship domain values deviate from the expectation. The observed accuracy improvements are thus not entirely driven by a causal relationship between a predicted trajectory and the trajectories of nearby ships. This finding underscores the model’s explanatory capabilities through its intrinsically interpretable design. 
		Future work will focus on utilizing the architecture for counterfactual analysis and on the incorporation of more sophisticated attention mechanisms.
	\end{abstract}


\section{Introduction}
Inland shipping stands out as an indispensable mode of transportation with considerable unexploited potential. As a cost- and energy-efficient alternative, it offers a promising pathway toward sustainable and competitive freight movement compared to other transport modes (\cite{Calderon-Rivera2024-lb}).
With a growing demand, automation of inland shipping is gaining increasing relevance. Trajectory prediction is a fundamental component of autonomous shipping systems, necessary to address the challenges of dense traffic environments, support collision avoidance, facilitate the safe coexistence of autonomous and human-operated vessels, and for an optimized traffic flow. Predicting inland vessel trajectories presents unique challenges compared to the maritime domain. 
Besides aspects of river geometry, fluid dynamics, and waterway regulations that impact ship trajectories in general, the accurate handling of ship-to-ship interactions constitutes a persistent difficulty regarding both the development of adequate models and their appropriate evaluation. Let alone the complicacy of model comparison in interaction-agnostic cases due to differences in the datasets, preprocessing, hyperparameter optimization and simplifying assumptions (\cite{Slaughter.2024}), especially the incorporation of interaction awareness raises questions regarding its evaluation. Specifically, it remains challenging to discern whether
\begin{itemize}
    \item{} improved predictions can be attributed to the interaction mechanism, \label{q:interaction_mechanism}
    \item{} these mechanisms inadvertently affect predictions in scenarios where interaction awareness is not essential. \label{q:nonessential_interactions}
\end{itemize}
Addressing these issues, an adaptation of the multi-ship trajectory prediction approach introduced in \cite{Sekhon.2020}, where interaction awareness is realized using a learnable ship domain, is proposed. Adjustments to the inland shipping context are made and limitations in the model architecture regarding the weighting of surrounding ships are revealed and addressed. To address \ref{q:interaction_mechanism} the ship domain parameters are used to indicate the extent to which the model has learned to regard an encounter type or if it has learned to ignore it entirely. To address \ref{q:nonessential_interactions} an architecture variant is proposed that separates the trajectory prediction into an interaction-agnostic and a interaction-aware path. This allows the model to predict interactionless trajectories without the involvement of the interaction attending component.
The remainder of this contribution is organized as follows. After situating the study within the broader research context in Section \ref{sec:related}, a problem definition and description of the proposed model variants are provided in Section \ref{sec:method}. The experimental results are presented in Section \ref{sec:results}, and the performance of the models and insights from the ship domain parameters are discussed. In Section \ref{sec:conclusion}, the paper concludes by highlighting the contributions and suggesting directions for future research.

\section{Related work}\label{sec:related}

Deep learning-based ship trajectory prediction methods can be classified as interaction-agnostic and interaction-aware. While most approaches are developed using maritime AIS datasets, some specifically address the inland shipping case (e.g., \cite{You.2020,Dijt2020,Donandt.2023}). The disparity between maritime and inland-specific approaches is more noticeable when interaction is considered. A few studies have evaluated models on both maritime and inland AIS datasets (\cite{Feng.2022,Jiang2024,Liu.2024}). Interaction-aware STP methods tailored specifically to the inland shipping contexts are still rare. One example is the work of \cite{Donandt.2023SMC} where the surrounding traffic during observation is represented in a social tensor (\cite{Alahi2016}) which is jointly processed with the target vessel trajectory features. Predictions, however, are only generated for the target ship without considering the potential future behavior of the surrounding ships. To the best of the authors' knowledge, inland-specific multi-ship trajectory prediction (MSTP) models, which jointly predict the trajectories of all vessels in a scene, do not yet exist. 

Some MSTP approaches use the concept of a ship domain, which is defined as a two-dimensional area around a vessel that must remain clear of intrusion by surrounding vessels or obstacles (\cite{Liu.2023}).~ \cite{Feng.2022} propose a combination of the Social-STGCNN (\cite{Mohamed2020}) for trajectory prediction with Model Predictive Control for the adjustment of the predictions to comply with kinematic constraints. For training of the GNN-based trajectory prediction model, a sampling approach incorporating both positive samples (i.e., ground truth positions) and negative samples (i.e., ``fake'' positions) is used. The bumper model, defining a ship domain only depending on the ship length, is applied to generate negative samples by selecting trajectory points that fall within the bumper model zone of a nearby ship. ~\cite{Liu.2023} employ an LSTM encoder-decoder model for MSTP. The Quaternion Ship Domain (QSD) (\cite{Wang2010}) 
, a ship domain depending on the vessel’s speed, length, advance, and tactical diameter, is used to identify influential vessels. The hidden states of nearby vessels falling within the QSD of a given target are aggregated in a social pooling layer using the distance to the target and the repulsion force magnitude (\cite{Tordeux2016}). The resulting interactive state is combined with a representation of the target and fed to the LSTM. 

In contrast to the previous approaches that rely on a predefined ship domain, \cite{Sekhon.2020} propose determining the dynamic extensions of the ship domain during the training of an LSTM encoder-decoder-based MSTP model with spatial and temporal attention mechanisms. During training, a parameter matrix is optimized, in which the cell indices represent intervals for the relative heading and relative bearing between each ship and its neighbors, and the cell entries represent the ship domain value corresponding to these intervals. The weights used to combine the hidden states of target and neighboring ships in the spatial attention mechanism are obtained by comparing the ship domain values to the actual distances. For smaller ship domain values, the weights are set to 0, implying the exclusion of the corresponding vessels' hidden states in the fused hidden state calculation; otherwise, the weights are equal to the difference between the distances and the respective ship domain values. 
In this study, the approach of \cite{Sekhon.2020} is adapted to the inland domain. Instead of fusing the hidden states of all vessels, however, a clear separation of the representations for the target vessel's state and the interactive state as in \cite{Liu.2023} is realized. Different from the latter approach, the fusion of these separate states is done on the hidden state level and the learnt ship domain parameter values are used to combine the surrounding vessels' hidden states. Furthermore, different to \cite{Sekhon.2020}, the selection of relevant ships through the ship domain parameter matrix and the subsequent hidden state fusion are clearly separated into distinct and independently exchangeable components, facilitating future XAI approaches such as counterfactual analysis.
\section{Methodology}\label{sec:method}
The task of multi-ship trajectory prediction can be formally described as a function from $X^{T_{obs} \times |V|}$ to $Y^{T_{pred} \times |V|}$, where $X$ and $Y$ denote the observational and predictive feature spaces, $T_{obs}$ and $T_{pred}$ denote the observation and prediction time steps (in minutes), and $V$ the set of ships present in a traffic situation. Specifically, $X\ms\ni\ms{x}_i^{t}\ms=\ms(k_i^{t},f_i^{t},\Delta{k}_i^{t},\Delta{f}_i^{t},c_i^{t},\hat{c}_i^{t})$ describes the position of an inland vessel $i\in V$ at a given observation time step $t \in \{1,\cdots,T_{obs}\}$ by the waterway kilometer (wkm) $k_i^{t}$ (as the longitudinal position), the offset from the fairway center  $f_i^{t}$ (as the lateral position), and both positional changes between two subsequent prediction time steps as $\Delta{k}_i^{t}=k_i^{t}-k_i^{t-1}$ and $\Delta{f_i^t} = f_i^{t} - f_i^{t-1}$.
\begin{figure}[h!]
    \centering 
    \includegraphics[width=0.45\textwidth,trim=1cm 0.4cm 0.9cm 0.3cm,clip]{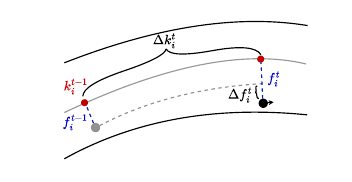}
    \caption{Illustration of positional features wkm as $k_i^t$, offset from the fairway axis as $f_i^t$, and their changes with respect to the previous time step $\Delta k_i^t$ and  $\Delta f_i^t$ .}
    \label{fig:positional_features} 
\end{figure}
The curvature of the waterway axis $c_i^{t}$ and the direction of the curve $\hat{c}_i^{t}$ are included in consideration of a curved environment being transformed into Cartesian coordinates (cf. \cite{Donandt.2023}). Instead of directly predicting the vessel's positions, $y_i^{t} \in Y$ consists of the positional distances  $\Delta{k}_i^{t+T_{obs}}$ and $\Delta{f}_i^{t+T_{obs}}$. In this study, the observational and predictive time horizons are equal, hence $T_{obs} = T_{pred}=:T$.

The function from $X$ to $Y$ is realized as an LSTM-based Encoder-Decoder model with embedded global Dot-Product-Luong-Attention (\cite{luong2015}), and different variants are proposed herein. The Encoder-Attention-Decoder-Attention model (EA-DA) updates the hidden state of each vessel $i$ at time step $t$ during observation in its encoder as 
\begin{equation}
    h_i^{t} = EncLSTM(x_i^{t}, \alpha(\omega_i(h^{t-1}))), \label{eq:enc}
\end{equation}
where $h^{t-1}$ are the hidden states of all vessels in the traffic situation at the previous time step, $h^0$ is a zero-valued tensor, $\omega_i$ is a weighting function from vessel $i$'s perspective, and $\alpha$ an attention function. These functions are given by
\begin{align}
    \omega_i(h^t)&=(w_{ij}^th_j^t)_{j \in N_i}, \label{formula:w} \\
    \alpha(\omega_i(h^t))&=\sum_{\substack{j\in N_i}} (\omega_i(h^t))_j, \label{formula:a} 
\end{align}
where $N_i=V$. The weight $w_{ij}$ is obtained as 
\begin{align}
    & w_{ij}^t = \mathrm{max}\left(S\left(\Gamma_{ij}^t, \Theta_{ij}^t, \Phi_{ij}^t\right) - \Delta_{ij}^t,0\right), \label{eq:s}
\end{align}
using a learnable ship domain parameter tensor $S$. For a triple of discretized, inland-specific ship-to-ship relation values, $\Gamma_{ij}^t$, $\Theta_{ij}^t$, and $\Phi_{ij}^t$, $S\left(\Gamma_{ij}^t, \Theta_{ij}^t, \Phi_{ij}^t\right)$ returns a domain parameter in the unit of wkm, which is then compared to the distance between $i$ and $j$, $\Delta_{ij}^t =|k_i^t-k_j^t|$. 

\begin{figure}[h!]
    \centering 
    \includegraphics[width=0.45\textwidth,trim= 0.8cm 0.2cm 0.8cm 0.4cm,clip]{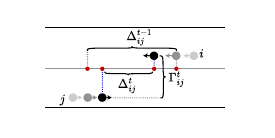}
    \caption{Illustration of ship-to-ship relations for opposing ships ($\Theta \ms=\ms-1$). The change in longitudinal distance $\Phi_{ij}^t$ is given by $\Delta_{ij}^t - \Delta_{ij}^{t-1}$.}
    \label{fig:gamma_theta_phi} 
\end{figure}
Here, $\Gamma_{ij}^t$, $\Theta_{ij}^t$, and $\Phi_{ij}^t$ are obtained from the lateral distance $|f_i^t-f_j^t|$, the relative direction of movement, $sign(\Delta{k}_i^t)\times sign(\Delta{k}_j^t)$, and from the distance change rate, 
$\Delta_{ij}^t - \Delta_{ij}^{t-1}$, respectively. An illustration for the calculation of the ship-to-ship relation values is given in Fig. \ref{fig:gamma_theta_phi}. In Tab. \ref{tab:gamma_theta_phi_buckets}, the categories and intervals used for discretization resulting in a $3 \times 4 \times 4$-dimensional $S$ are detailed. Note that the trained \( S \) solely depends on combinations of \( \Gamma \), \( \Theta \), and \( \Phi \) values, abstracting from any specifically observed trajectories \( x_i \) and \( x_j \). Thus, \( S(\Gamma, \Theta, \Phi) \) can be interpreted as an awareness range for the encounter type constituted by \( \Gamma, \Theta, \Phi \). Particularly, a \( S(\Gamma, \Theta, \Phi) \) close to 0 indicates complete irrelevance of the corresponding encounter type, while a high value suggests relevance for that type at greater distances. 

\begin{figure}[t!]
	\centering
    \begin{subfigure}[t]{0.5\textwidth}
    \centering 
    \includegraphics[trim=0.985cm 0.35cm 0.715cm 0.5cm,clip]{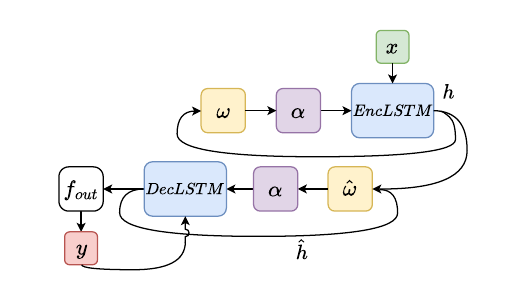}
    \caption{EA-DA}
    \label{fig:component_diagram_sekhon}
    \end{subfigure}%
     \vspace{10pt}
    \begin{subfigure}[t]{0.5\textwidth}
    \centering 
    \includegraphics[trim=0.985cm 0.35cm 0.715cm 0.5cm,clip]{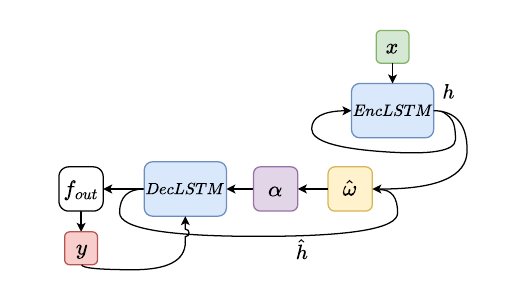}
    \caption{E-DA}
    \label{fig:comp_diag_simple} 
     \vspace{15pt}    
\end{subfigure}
     \vspace{10pt}
    \begin{subfigure}[t]{0.5\textwidth}
        \includegraphics[trim=0.985cm 0.35cm 0.715cm 0.5cm,clip]{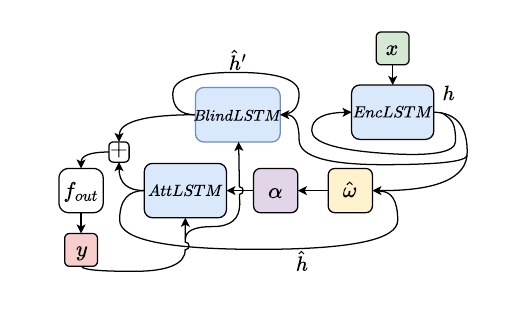}            \caption{E-DDA}\label{fig:comp_diag_dual} 
    \end{subfigure}
    \caption{Model variants}
\end{figure}

The EA-DA decoder LSTM updates its hidden state $\hat{h}$ similar to Eq. \eqref{eq:enc} and the output at each prediction time step $t$ is obtained as 
\begin{align}
    \hat{h}_i^{t} & = DecLSTM(y_i^{t-1}, \alpha(\hat{\omega}_i(\hat{h}^{t-1}))) \\
    y_i^t & = f_{out}(\hat{h}_i^{t}),
\end{align}
where $f_{out}$ is a Feed Forward Network, $\hat{h}^0=h^{T}$,  $y^0_i\ms=\ms(\Delta \ms{k_i^{T}},\Delta \ms{f_i^{T}})$, and $\hat{\omega}_i$ is defined analog to Eq. \eqref{formula:w} as
\begin{equation}
    \hat{\omega}_i(\hat{h}^t)=(w_{ij}^{t+T}\hat{h}_j^t)_{j \in N_i}.\label{formula:w_hat}
\end{equation}
Note that LSTM cell states are not included here for the purpose of simplicity. In Fig. \ref{fig:component_diagram_sekhon}, the architecture of EA-DA is depicted. 

Unlike \cite{Sekhon.2020}, where weighting and attending are integrated in a single concept of spatial attention, this model considers them as two separate aspects:
\begin{itemize} 
    \item{} The weighting function $\omega$ is expected to prioritize and preselect hidden states deemed relevant with respect to the current ship-to-ship relation values $\Gamma$, $\Theta$, and $\Phi$. \label{q:weighting_func}
    \item{} The attention function $\alpha$ is expected to deduce an interaction-aware hidden state 
    from the preselected ones in \ref{q:weighting_func}.
\end{itemize}
\begin{table}[h!]
	\centering
    \caption{Discretisation of ship-to-ship relation values. \label{tab:gamma_theta_phi_buckets}}
    {\tabcolsep10pt
\begin{tabular}{@{}ccc@{}}
        \hline 
        \textbf{Relative} & \textbf{Distance} & \textbf{Lateral} \\ 
         \textbf{direction $\Theta$} & \textbf{change $\Phi$} & \textbf{distance $\Gamma$} \\ 
          \textbf{} & \textbf{(wkm/min)} & \textbf{(m)} \\ 
         \hline
        -1 (opposing)                     & $< -0.2$              & $0 - 10$ \\ \hline
        1 (aligned)                      & $-0.2$ to $-0.05$       & $10 - 20$  \\ \hline
        0 (stationary)                   & $-0.05$ to $+0.05$ & $20 - 40$   \\ \hline
        --                           & $> 0.05$                & $> 40$      \\ \hline 
    \end{tabular}}
\end{table}
Two variants of the EA-DA are developed: 
the Encoder-Decoder-Attention model (E-DA) (Fig. \ref{fig:comp_diag_simple}) and the Encoder-Dual-Decoder-Attention model (E-DDA) (Fig. \ref{fig:comp_diag_dual}). Compared to EA-DA, E-DA only uses the weighting and attention complex in the decoder LSTM to reduce the impact of the attention component to one specific point in the model. This reduction aims to facilitate future analysis towards the causality incorporated in and the added value given by such a component. 
The E-DDA model is an E-DA where the decoder is split into an interaction-blind and an attention decoder, $BlindLSTM$ and $AttLSTM$. The prediction for vessel $i$ at time step $t$ is obtained as 
\begin{align}        
    y_i^t = f_{out}\left(\substack{BlindLSTM(y_i^{t-1}, \hat{h}'^{t-1}_i),\\ 
            AttLSTM(y_i^{t-1}, \alpha(\hat{\omega}_i(\hat{h}^{t-1}))}\right),
\end{align}
where $\hat{h}'$ are the hidden states of $BlindLSTM$ and $N_i=V\setminus\{i\}$ with respect to Formula \eqref{formula:a} and \eqref{formula:w_hat}. Thus, the $BlindLSTM$ only processed vessel $i$'s own hidden states, whereas $AttLSTM$ processes the weighted and attended  hidden states of the foreign vessels only. This differs from EA-DA and E-DA, where $w_{ii}$ is calculated and therefore the ship-to-ship relation of vessel $i$ to itself is represented in a $(\Gamma,\Theta,\Phi)$-combination (see Tab. \ref{tab:gamma_theta_phi_buckets}). 

In the following the proposed models are compared by their final displacement error at varying prediction horizons $t \in \{1,...,T\}$ defined as 
\begin{align}
    FDE_t = \frac{\sum_{i \in I} d\left(y^t_i, \hat{y}^t_i\right)}{|I|}, 
\end{align}
where $d$ is the Euclidean distance, and $I$ denotes the validation set consisting of prediction-ground-truth tuples $(y^t,\hat{y}^t)$. Note, that not all vessels have predictions up to time step $T$ as some exit the considered waterway area before completing $T$ time steps. Thus, more prediction-ground truth tuples exist for the shorter prediction horizons than for the longer ones. After the performance comparison, the soundness of the learned ship domain parameters with respect to the ship-to-ship encounters incorporated by realisations of $\Gamma$, $\Theta$, and $\Phi$ is investigated.

\section{Results}\label{sec:results}
\begin{figure}[h!]
    \centering 
    \includegraphics[trim=0.1cm 0.0cm 0.0cm 0.0cm,clip,width=0.47\textwidth]{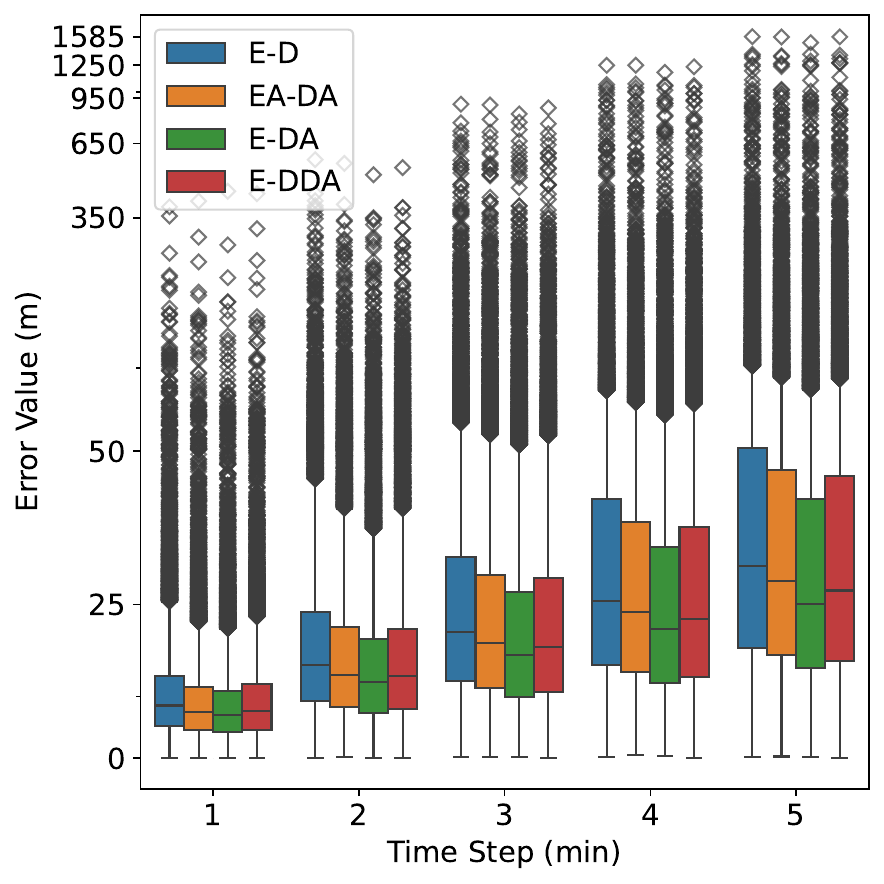}
    \caption{$FDE_t$ per model and time steps. The error values are linearly scaled from 0 to 50 and $log_{10}$-scaled beyond that range. Outliers are shown as diamonds.}
    \label{fig:fde_boxplots_per_ts} 
\end{figure}

An AIS dataset for a selected section of the Rhine (Rhine kilometers 595 - 611) covering more than 3 years (January 2021 - April 2024) is used in this study. It consists of regularly sampled time series (1 min time step size), referred to as situations. A situation contains all vessel trajectories present in the selected river section. The sizes of the training, validation, and test datasets (as number of situations and total number of trajectories) are given in Tab. \ref{tab:dataset_sizes}.
\begin{table}[h!]
	\centering
    \caption{Sizes of datasets in number of trajectories and situations.}
    {\tabcolsep10pt
    \begin{tabular}{@{}lccc@{}}
        \hline
                   & \textbf{Training} & \textbf{Validation} & \textbf{Test} \\ \hline
        Situations & 15,920    & 2,079  & 2,055 \\ \hline
        Total & \multirow{2}{*}{159,673}  & \multirow{2}{*}{21,178} & \multirow{2}{*}{20,758} \\ 
        trajectories & & & \\ 
        \hline
    \end{tabular}}
    \label{tab:dataset_sizes}
\end{table}
With a temporal horizon of $T = 5$ for observation and prediction and the ship domain parameter tensor initialized as $S(\cdot,\cdot,\cdot)=0.1$, the models are trained for 750 epochs. Since the hidden state size of 6 proposed by \cite{Sekhon.2020} performs poorly, a significantly higher one (300) is chosen. For all model variant, the distribution of each $FDE_t$ is depicted in Fig. \ref{fig:fde_boxplots_per_ts}. As a benchmark model, a further simplified variant of E-DA is adopted by removing the attention component from the decoder, resulting in a standard Encoder-Decoder model (E-D) with incorporated Luong-Attention.
According to these boxplots, E-DA achieves the lowest errors and E-DDA performs better than EA-DA. The mean, median, and standard deviation ($\sigma$) of $FDE_5$ (see Tab. \ref{tab:t5_fdes}) indicate the same ranking of the model variants. The interaction-agnostic benchmark model performs worst across all metrics.
\begin{table}[h!]
	\centering
    \caption{Statistic values of $FDE_5$.\label{tab:ship_statistics}}
    {\tabcolsep10pt
    \begin{tabular}{@{}lcccc@{}}
        \hline
        \textbf{Model}    & \textbf{E-D} & \textbf{EA-DA} & \textbf{E-DA} & \textbf{E-DDA} \\ \hline
        \textbf{Mean}     & 45.1 & 41.9 & 38.4 & 40.9 \\
        \textbf{Median}   & 31.2 & 28.8 & 25.1 & 27.3 \\ 
        \textbf{$\sigma$} & 64.7 & 61.9 & 60.1 & 61.0 \\ \hline
        \vspace{-20pt}
    \end{tabular}}
    \label{tab:t5_fdes}
\end{table}
A visual analysis of the prediction results with extreme outliers (more than 300 m) shows that they can mostly be attributed to very specific events, especially ship leaving a port or a mooring. These cases are not expected to be predicted well and also not a concern of ship-to-ship interaction. Overall, the proposed models do satisfactorily predict sound trajectories.  
\begin{figure}[h!]
    \centering
    \begin{subfigure}[t]{0.47\textwidth}
        \centering 
        \includegraphics[trim=0.1cm 0.0cm 0.2cm 0.0cm,clip,width=1\textwidth]{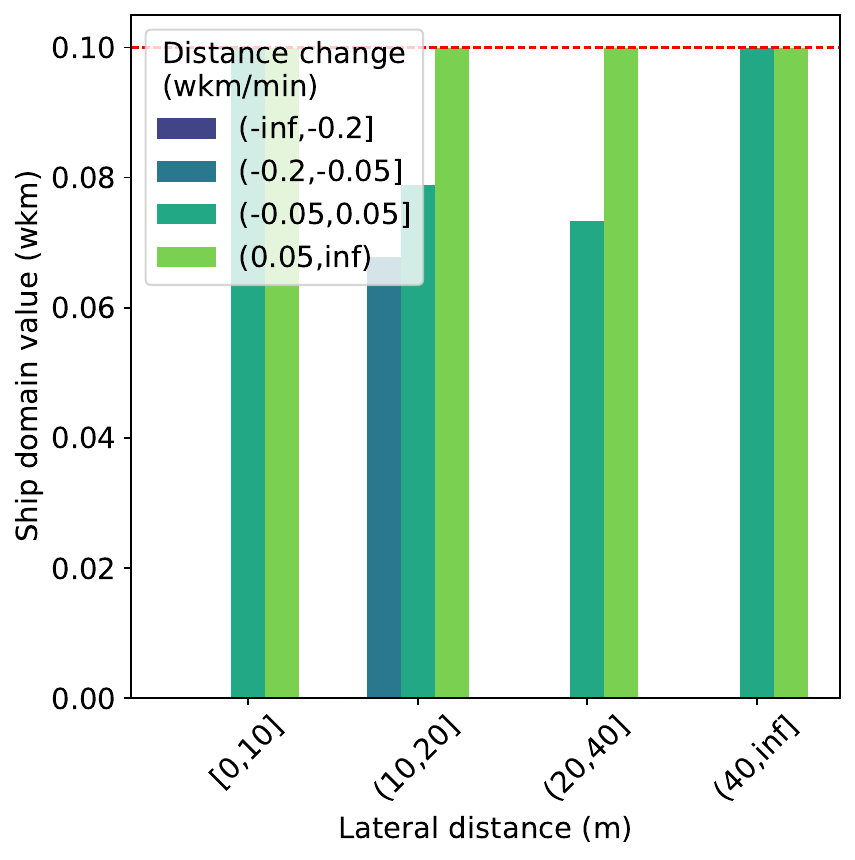}
        \caption{E-DA}
        \label{fig:eda_opposing}
    \end{subfigure}%
    \hfill
    \vspace{6pt}
    \begin{subfigure}[t]{0.47\textwidth}
        \centering 
        \includegraphics[trim=0.1cm 0.0cm 0.2cm 0.0cm,clip,width=1\textwidth]{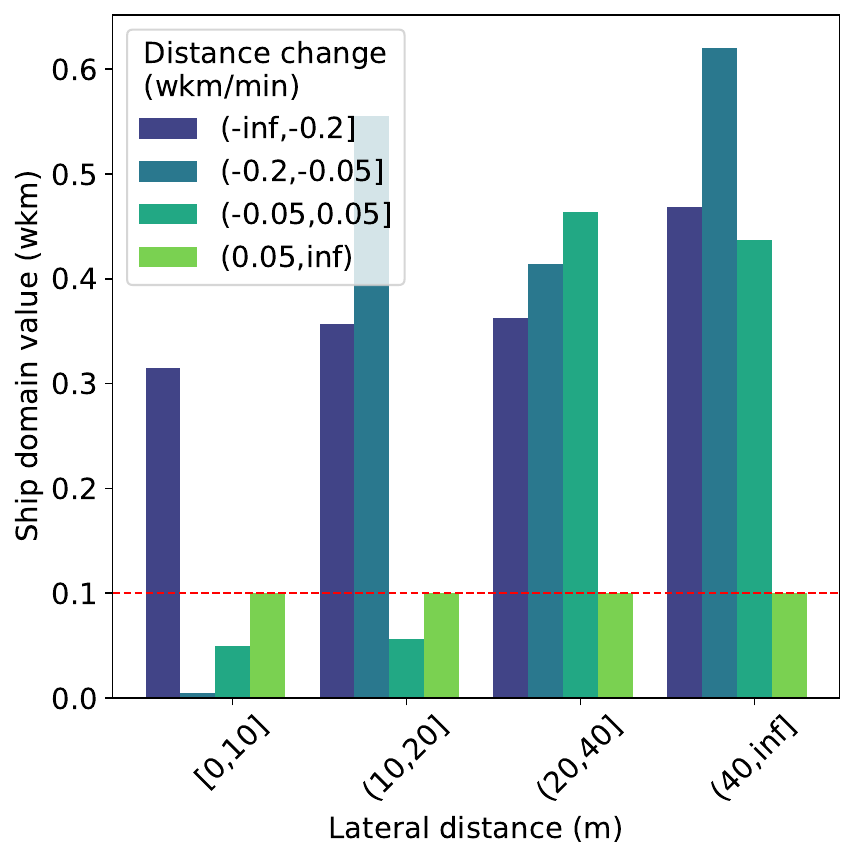}
        \caption{E-DDA}
        \label{fig:edda_opposing} 
    \end{subfigure}
    \hfill
    \caption{Ship domain values for vessels that are moving in opposite directions. The red dashed line indicates the initial ship domain value.}
\end{figure}
Though having the lowest $FDE_t$ values, the learned ship domain of E-DA, as depicted in Fig. \ref{fig:eda_opposing} for vessels moving in opposing direction, indicates that this predictive quality does not rely on its ability to attend and interpret ship-to-ship interaction for most encounter types. The model has not increased its ship domain values for any opposing ships, on the contrary decreasing it for any opposing ship-to-ship relation with a negative distance change. Hence, when predicting vessel $i$'s next position, any opposing ship $j$ with a distance of more than 0.1 wkm is deemed irrelevant for the prediction of i'th next position as the corresponding weight obtained from Formula \eqref{eq:s} is 0. For EA-DA, the learned ship domain parameter is similar. In contrast, the E-DDA model's learned ship domain values tend towards the expected behavior: the initial ship domain values of 0.1 wkm have predominantly been increased during training for  encounters with a negative distance change of opposing ships (see Fig. \ref{fig:edda_opposing}). Therefore, an opposing and approaching ship is indeed taken into account for the prediction of ship $i$. The difference in ship domain values between the EA-DA and E-DA models on one side and the E-DDA model on the other, observable also for ships moving in the same direction, is likely to stem from the indistinction of target and neighboring ships in the weighting function of the EA-DA and E-DA variants. This issue is addressed in the E-DDA variant by explicitly excluding the target ship from the weighting mechanism. While this adjustment results in a seemingly more plausible weighting, the E-DDA model shows a trend to attribute higher importance to surrounding vessels that have a higher lateral distance to the target. This behavior appears counterintuitive, as ships at small lateral distance are generally expected to have greater relevance to a target ship's future trajectory. Possible explanations are that either ships in close proximity have already set their course or that ships near the fairway boundary - after giving way to a passing ship - try to return to their typical position closer to the center of the fairway. These hypotheses, however, require further analysis of the cases corresponding to the respective encounter types to be confirmed.
\section{Conclusion and Outlook}\label{sec:conclusion}
This paper proposed three variants of interaction-aware multi-ship trajectory prediction tailored to inland waterway shipping. A design that separates the selection of relevant ships through a trainable ship domain parameter from the attention mechanism for hidden state fusion was explored, offering distinct and independently exchangeable components. The evaluation of ship domain parameters revealed that models with an overall well performance do not derive that quality from an effective implementation of interaction awareness. This underscores the need for evaluation methods beyond displacement error metrics to avoid erroneous conclusions about the role of a presumed interaction awareness in prediction quality. Future work will focus on comparing alternative attention functions for the hidden state fusion, on improving the encounter type definition, and leveraging the weighting function as a basis for counterfactual analysis.
\paragraph{Acknowledgement}
The authors thank the German Federal Waterways Engineering and Research Institute for the provision of AIS and river-specific data, the trip splitting and kilometerization algorithms, and the training infrastructure.

\bibliographystyle{unsrt}
\bibliography{References}

\end{document}